\documentclass[10pt,journal,compsoc]{IEEEtran}

\usepackage{algorithm}
\usepackage{algorithmic}
\usepackage{balance}
\usepackage{multirow}
\usepackage{hyperref}

\ifCLASSOPTIONcompsoc
  \usepackage[nocompress]{cite}
\else
  \usepackage{cite}
\fi
\usepackage{graphicx}
\ifCLASSINFOpdf
\else
\fi
\usepackage{amsmath}
\usepackage{xcolor}
\usepackage{tabularx}
\usepackage{graphicx}
\usepackage{adjustbox}

\begin{document}
%
\title{Few-shot Learning in Emotion Recognition of Spontaneous Speech Using a Siamese Neural Network with Adaptive Sample Pair Formation
}
%
%
%
%

\author{Kexin~Feng,~\IEEEmembership{Student~Member,~IEEE,} and~Theodora~Chaspari,~\IEEEmembership{Member,~IEEE}
\IEEEcompsocitemizethanks{\IEEEcompsocthanksitem K. Feng and T. Chaspari are with the HUman Bio-Behavioral Signals (HUBBS) Laboratory in the Department of Computer Science \& Engineering at Texas A\&M University, College Station, TX, 77843.
\protect\\
E-mail: kexin@tamu.edu and chaspari@tamu.edu\\
The code implemented in this paper and related documentation can be found at: \url{https://github.com/HUBBS-Lab-TAMU/pytorch-MeL_S_ASPF}.}
}

%
%

\markboth{ IEEE Transactions on Affective Computing}%
{Shell \MakeLowercase{\textit{et al.}}: Bare Advanced Demo of IEEEtran.cls for IEEE Computer Society Journals}
%



\IEEEtitleabstractindextext{%
\begin{abstract}
Speech-based machine learning (ML) has been heralded as a promising solution for tracking prosodic and spectrotemporal patterns in real-life that are indicative of emotional changes, providing a valuable window into one's cognitive and mental state. Yet, the scarcity of labelled data in ambulatory studies prevents the reliable training of ML models, which usually rely on "data-hungry" distribution-based learning. Leveraging the abundance of labelled speech data from acted emotions, this paper proposes a few-shot learning approach for automatically recognizing emotion in spontaneous speech from a small number of labelled samples. Few-shot learning is implemented via a metric learning approach through a siamese neural network, which models the relative distance between samples rather than relying on learning absolute patterns of the corresponding distributions of each emotion. Results indicate the feasibility of the proposed metric learning in recognizing emotions from spontaneous speech in four datasets, even with a small amount of labelled samples. They further demonstrate superior performance of the proposed metric learning compared to commonly used adaptation methods, including network fine-tuning and adversarial learning. Findings from this work provide a foundation for the ambulatory tracking of human emotion in spontaneous speech contributing to the real-life assessment of mental health degradation.
\end{abstract}
\vspace{-5pt}
\begin{IEEEkeywords}
Emotion recognition, scripted/spontaneous speech, few-shot learning, metric learning, siamese neural network
\end{IEEEkeywords}}

\maketitle

\IEEEdisplaynontitleabstractindextext

\IEEEpeerreviewmaketitle

\ifCLASSOPTIONcompsoc
\IEEEraisesectionheading{\section{Introduction}\label{sec:introduction}}
\else
\section{Introduction}\vspace{-5pt}
\label{sec:introduction}
\fi
\IEEEPARstart{E}{motion} tracking has been investigated as an approach to help individuals remain in psychologically healthy and to assist with mental diseases~\cite{compare2014emotional}. Speech-based ambulatory monitoring is a promising method to longitudinal emotion tracking: prosodic and spectrotemporal patterns of speech reflect changes in muscle tension of the articulatory system, which are indicative of one's emotions~\cite{liebenthal2005neural}. Supplemented with artificial intelligence (AI), speech has the potential to serve as a valuable biomarker for tracking emotions and triggering early mental health intervention mechanisms~\cite{marchi2016real}.


Despite the potential of ambulatory speech monitoring for tracking human emotion, the reliable annotation of spontaneous emotional speech (i.e., speech elicited in realistic conditions without planning), usually conducted via self-reports or third-party evaluators, is an inherently challenging task. Self-reporting is subjective and potentially biased by social, cultural, and psychological factors~\cite{stone1999science}, while third-party annotation can be erroneous and time-consuming~\cite{metallinou2013annotation}. Due to these limitations, audio samples in many speech-based emotion datasets are collected through acted elicitation methods relying on individuals who engender a target emotion while uttering pre-determined linguistic contents, also known as scripted speech~\cite{haamer2017review}. Despite the fact that these methods tend to overlook subtle expression details, they provide ample data, based on which machine learning (ML) methodologies can recognize emotions~\cite{chenchah2014speech}.

Transfer learning refers to leveraging knowledge from ample training examples from one domain to learn robust data representations for another (potentially related) domain~\cite{pan2009survey}. Various transfer learning algorithms have been proposed for speech emotion recognition, including adaptive support vector machines, neural network fine-tuning, progressive neural networks, and adversarial learning~\cite{gideon2017progressive, li2019exploring, song2017transfer, zhou2019transferable}. Despite the promising results, these require a relatively large number of samples from the target domain to achieve promising performance. Few-shot learning has been proposed as an alternative to fully-supervised transfer learning, since it accounts for the potential shortage of labelled samples in the target domain. A promising approach in few-shot learning relies on {\it Metric Learning (MeL)}. MeL learns a transferable distance-based embedding that models the relative distance between classes~\cite{hilliard2018few}. In this way, it is easier to classify samples based on the new embedding rather than the original space. MeL has been explored for audio classification~\cite{zhang2019few, wang2020few}, speaker recognition~\cite{li2020automatic}, and image-based emotion and facial expression recognition~\cite{zhan2019zero,wang2020learning}. To the best of our knowledge, MeL has not been examined in speech-based recognition of spontaneous emotion.

We propose a MeL approach that transfers knowledge from scripted to spontaneous speech by learning emotion-specific speech embeddings with small supervision from the target domain. MeL conducts pairwise comparisons of samples between emotional classes and is implemented with a siamese neural network (SNN). Beyond the pairwise sample comparison, we impose an additional supervision to MeL that allows to directly obtain an emotional class output for a test sample, referred to as {\it Metric Learning with Supervision (MeL-S)}. Finally, we address issues related to training convergence that yield by the random pair formation in MeL and MeL-S through an adaptive procedure. The proposed {\it Metric Learning with Supervision and Adaptive Sample Pair Formation (MeL-S-ASPF)} iteratively trains a SNN using an adaptive selection likelihood of the input samples, assigning higher probability to consistently misclassified samples. The proposed few-shot learning is compared to in-domain learning, out-of-domain learning, as well as transfer learning implemented with feedforward neural network (FNN) fine-tuning and adversarial learning. Results indicate the ability of MeL-S to effectively transfer knowledge between acted and spontaneous speech relying only on a small number of labelled samples from the target domain (e.g., 70\% unweighted average recall using 2-3 labelled samples per class on a 3-way classification task). Our findings further suggest that the proposed MeL-S-ASPF provides advantages compared to random sample pair formation of MeL and MeL-S, which are especially beneficial when using a small number of labelled samples.

\vspace{-8pt}
\section{Previous Work}
\vspace{-5pt}
Previously proposed transfer learning methods for speech-based emotion recognition include both supervised and unsupervised approaches. The first require labeled data from the target domain, in contrast to the latter. In terms of supervised learning, prior work has explored the effect of supervised domain adaptation in cross-corpus experiments using adaptive and incremental support vector machines~\cite{abdelwahab2015supervised}, as well as auto-enconder architectures that learn a transferable low-dimensional feature space from the original features~\cite{deng2013sparse,sahu2017adversarial}. Progressive neural networks have been also proposed as an alternative method to transfer knowledge between domains without forgetting the learned embeddings of the source domain~\cite{gideon2017progressive,li2019exploring}. In terms of unsupervised transfer learning, adversarial and generative learning have been employed to tackle the distribution mismatch between emotional speech corpora~\cite{abdelwahab2018domain,gideon2019improving,chang2017learning}. Results demonstrate that even including unlabelled data from the target domain can yield considerable improvement compared to not including any target data. A detailed review of transfer learning methodologies with focus on speech-based emotion recognition can be found in~\cite{feng2020transfer}.

Few-shot learning aims to classify samples from a target domain using a small number of labelled examples from that domain. A promising few-shot learning approach relies on metric learning algorithms, which aim to learn a transferable feature embedding by optimizing a distance loss metric between the classes of interest, rather than learning the distribution of each class separately~\cite{lake2011one,sung2018learning}. Metric learning, implemented through SNNs~\cite{koch2015siamese} and relation networks~\cite{sung2018learning}, has been explored in person re-identification, speaker verification, and recommendation systems \cite{yu2017cross,tay2018latent,novoselov2018triplet}. SNNs have also shown promising results in in-domain emotion classification, not in the context of few-shot learning~\cite{lian2018speech,huang2018speech,sabri2018facial}. As part of our preliminary work, we have explored the feasibility of metric learning implemented with SNN to few-shot emotion recognition~\cite{feng2020a}.

The contributions of this paper are: (1) A novel formulation of MeL and MeL-S for transferring emotion-specific knowledge between a source and a target domain using a small number of labelled samples from the target domain. The proposed approach is likely to overcome limitations of conventional distribution-based learning due to its ability to model the relative distance between classes; (2) An iterative adaptive sample pair formation for SNN training, that promotes the selection of samples which are consistently misclassified by previous iterations. This has the potential to increase the robustness of metric learning compared to modeling the distance between random pairs of samples; and (3) An investigation of the feasibility of transferring knowledge between acted (source) and spontaneous (target) speech, providing a foundation into ways to leverage the large number of labelled samples from acted speech datasets in emotion tracking in spontaneous speech.

\vspace{-8pt}
\section{Data Description and Pre-Processing}
\label{sec:DataDescr}
\vspace{-5pt}
In this study, we use four datasets, which were selected due to the fact that they contain utterances spoken in English and categorical emotional labels. These include the Interactive Emotional Dyadic Motion Capture (IEMOCAP)~\cite{busso2008iemocap}, Ryerson Audio-Visual Database of Emotional Speech and Song (RAVDESS)~\cite{livingstone2018ryerson}, Crowd-sourced Emotional Multimodal Actors Dataset (CREMA-D)~\cite{cao2014crema}, and Audio-Visual Emotion Database (eNTERFACE'05)~\cite{martin2006enterface}. IEMOCAP~\cite{busso2008iemocap} consists of dyadic sessions between 10 actors, who are engaged in both scripted (IEMOCAP-SC) and spontaneous (IEMOCAP-SP) interactions. RAVDESS~\cite{livingstone2018ryerson} contains audio-visual recording from 24 actors, who uttered a list of scripted sentences with different emotions. CREMA-D~\cite{cao2014crema} includes speech data from 91 actor participants, who were asked to express scripted sentences with target emotions until approved by an expert. The eNTERFACE’05 dataset~\cite{martin2006enterface} includes 42 speakers, who were asked to utter scripted sentences in response to listening short stories that were used to elicit various target emotions. Here, we explore the three most common emotions present in these datasets: anger, happiness, and sadness. Scripted speech samples from IEMOCAP-SC (133 minutes), RAVDESS (36 minutes), CREMA-D (159 minutes), and eNTERFACE’05 (30 minutes) are used as the source data, while spontaneous samples from IEMOCAP-SP (92 minutes) comprise the target data. Few-shot learning experiments will be conducted so that the samples from target dataset are gradually made available in the learning process.  

A 64-dimensional speech descriptor is extracted using the openSMILE toolkit \cite{eyben2010opensmile} to capture prosodic and spectrotemporal variations relevant to human emotion~\cite{sudhakar2015analysis}. The features are extracted using the configuration file from the INTERSPEECH’09 Emotion Challenge~\cite{schuller2009interspeech} with default parameters of $25ms$ frame length and $10ms$ step length. The first 32 feature dimensions include the arithmetic mean and standard deviation of frame-based speech intensity, zero-crossing rate, voicing probability, fundamental frequency, and the first 12 Mel-frequency cepstral coefficients (MFCC). The last 32 dimensions include the first-order derivative of the above descriptors. Feature normalization is conducted using the scikit-learn \cite{sklearn_api} library within each dataset to initially mitigate potential domain differences. Each feature is standardized using the mean and standard deviation computed using the samples of each dataset. The scripted and spontaneous part of the IEMOCAP dataset is normalized separately to avoid potential information leaking. 


\section{Proposed Methodology}
We will first introduce the fundamentals of the SNN architecture (Section~\ref{Sec:SNN}) and then describe the proposed metric learning approaches (Sections~\ref{Sec:Siamese}-\ref{Sec:Siamese&supervised&adaptive}). We will further provide details on the experimental design (Section~\ref{sec:ExpDesign}), including the formulation of in-domain and out-of-domain learning (Section~\ref{subsubsec:InOutDomain}), the experimental setting of the metric learning approaches (Section~\ref{subsubsec:FewShot}), and the baseline approaches (Section~\ref{subsubsec:Baseline}).

\vspace{-8pt}
\subsection{Siamese Neural Network (SNN)}
\label{Sec:SNN}
\begin{figure}[!tb]
  \centering
  \includegraphics[width=1\linewidth, trim = 0cm 0cm 0cm 0cm, clip=true, scale=0.7]{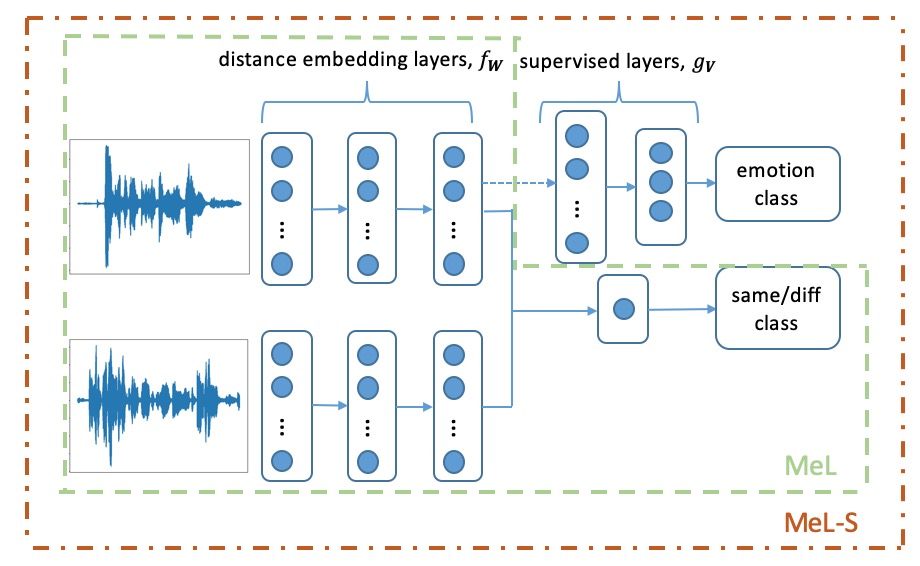}\vspace{-10pt}
 \caption{Visualization of the metric learning (MeL) and metric learning with supervision (MeL-S) models.}\vspace{-5pt}
\label{fig:model}
\end{figure}

\vspace{-5pt}
The Siamese neural network (SNN) is comprised of two input streams that compare a pair of input samples $(\mathbf{x_i},\mathbf{x_j})$ (Fig.~\ref{fig:model}). The hidden layers of the SNN learn a transformation $\mathbf{f}_\mathbf{W}$, parameterized by weights $\mathbf{W}$, that implements a similarity function between the two pairs of samples. The transformed input samples $\mathbf{f}_\mathbf{W}(\mathbf{x_i})$ and $\mathbf{f}_\mathbf{W}(\mathbf{x_j})$ are compared at the output through the distance function $d(\cdot,\cdot)$. The parameters $\mathbf{W}$ of the SNN are the same between the two input streams and are learned so that they minimize the following distance loss function:
\begin{equation}
    \begin{aligned}
    L_d(\mathbf{W})= &\sum_{c}{\sum_{\mathbf{x_i},\mathbf{x_j}\in\mathcal{X}_c}{d\left(\mathbf{f}_\mathbf{W}(\mathbf{x_i}),\mathbf{f}_\mathbf{W}(\mathbf{x_j})\right)}} -\\
    &\kappa\sum_{c\neq c'}{\sum_{\begin{matrix}\mathbf{x_i}\in\mathcal{X}_c \\ \mathbf{x_j}\in\mathcal{X}_{c'}\end{matrix}}{d\left(\mathbf{f}_\mathbf{W}(\mathbf{x_i}),\mathbf{f}_\mathbf{W}(\mathbf{x_j})\right)}}
\end{aligned}
\label{eq:1}
\end{equation}
where $\mathcal{X}_c$ is the set of data belonging to class $c$, $\mathcal{X}_{c'}$ is the set of data belonging to class $c'$ different than $c$, and $\kappa$ determines the trade-off between penalizing dissimilarity between samples belonging to the same class against similarity between samples belonging to different classes.

\vspace{-8pt}
\subsection{Metric Learning (MeL)}
\label{Sec:Siamese}
\vspace{-5pt}

We learn emotion-specific speech representations through a MeL approach implemented with SNNs. Our SNN is initially trained on samples of scripted speech (i.e., source) based on (\ref{eq:1}). Knowledge transfer is performed by fine-tuning the weights $W$ of the SNN using the small number of labelled samples from the spontaneous speech samples (i.e., target). Since the SNN is used to identify whether a pair of input samples belongs to the same emotion, it does yield explicit emotion labels. As a result, we obtain an emotion classification outcome for a test sample $\mathbf{x}$ by comparing its learned embedding to the embedding of the center $\bar{\mathbf{x}}_{\mathbf{c}}$ of the labelled samples from the target data for each emotional class $c$. The final emotion label is obtained by identifying the class whose center depicts the lowest distance to the test sample: $arg\min_c{d\left(\mathbf{f}_\mathbf{W}(\mathbf{x}),\mathbf{f}_\mathbf{W}(\bar{\mathbf{x}}_{\mathbf{c}})\right)}$.

\vspace{-8pt}
\subsection{Metric Learning with Supervision (MeL-S)}
\label{Sec:Siamese&supervised}
\vspace{-5pt}
We further impose an additional supervision constraint to the original SNN architecture, which allows us to directly obtain the emotional class outcome for a test sample (i.e., without having to compare the distance of the test sample with the center of samples from each class, as in Section~\ref{Sec:Siamese}). According to the MeL-S model, the transformed samples $\mathbf{f}_\mathbf{W}(\mathbf{x})$, where $\mathbf{W}$ is learned by the MeL approach through the loss function $L_d$, are fed into an additional set of fully-connected hidden layers $g_\mathbf{V}$, which learns a mapping between the transformed space $\mathbf{f}_\mathbf{W}(\mathbf{x})$ and the final class outcome $y$. The transformation $g_\mathbf{V}$ is implemented with a set of neural layers, but it could have been also implemented with any other linear or non-linear classification algorithm. The weights $\mathbf{V}$ are learned such that they minimize the cross-entropy loss of the emotion classification task:
\begin{equation}
    L_e(\mathbf{V})=-\sum_{\mathbf{x}\in\mathcal{X}, y\in\mathcal{Y}}{y\log{g_\mathbf{V}\left(\mathbf{f}_\mathbf{W}(\mathbf{x})\right)}}
\label{eq:2}
\end{equation}
where $\mathcal{X}$ is the set of input samples and $\mathcal{Y}$ is the set of labels in the training set.

The weights $\mathbf{W}$ are initially pre-trained using the MeL method on the source domain and refined using the Mel-S method on the target domain, such that $\{\mathbf{W}^*,\mathbf{V}^*\}=arg\min_\mathbf{V}\min_\mathbf{W}\left(L_e(\mathbf{V})+L_d(\mathbf{W})\right)$.

\vspace{-8pt}
\subsection{Metric Learning with Supervision and Adaptive Sample Pair Formation (MeL-S-ASPF)}
\label{Sec:Siamese&supervised&adaptive}
\begin{figure}[!tb]
  \centering
  \includegraphics[width=1\linewidth, trim = 0cm 0cm 0cm 0cm, clip=true, scale=0.7]{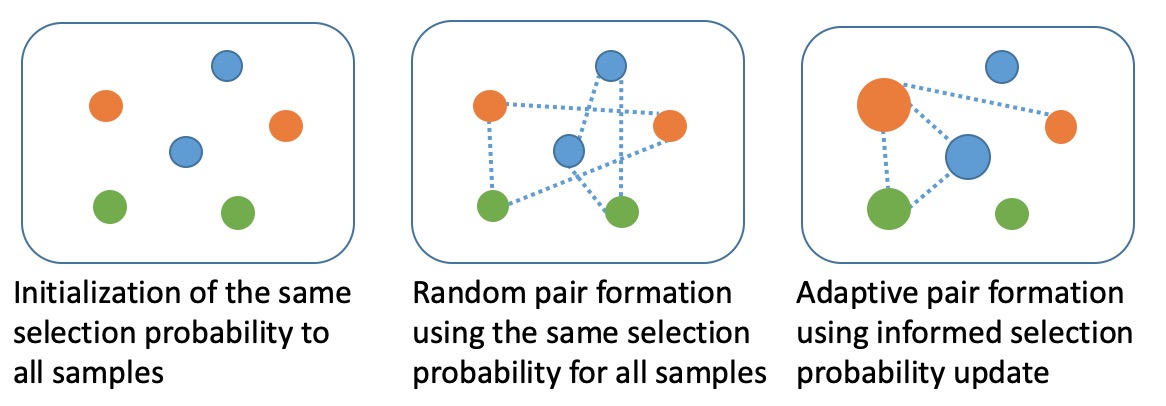}\vspace{-10pt}
 \caption{An example of the formation of sample pairs in metric learning with supervision and adaptive sample pair formation (MeL-S-ASPF).}
\label{fig:adaptive_weight_step}
\end{figure}

\vspace{-5pt}
The pairs of samples that serve as an input to the SNN in MeL and MeL-S (Sections~\ref{Sec:Siamese},~\ref{Sec:Siamese&supervised}) were randomly selected. In this way, convergence of the training process might require many iterations, since the algorithm has no information on how well specific samples are learned. To address this limitation, we propose the MeL-S-ASPF, an iterative learning process that alternates between training the SNN and assigning an importance factor to the input samples used for SNN training. The importance factor is assigned such that higher importance is given to samples that are not adequately learned by the SNN during previous iterations. Samples with higher importance have a higher chance to get selected in the next training iteration of the SNN (Fig.~\ref{fig:adaptive_weight_step}). According to MeL-S-ASPF, each sample of the training set $\mathbf{x}\in\mathcal{X}$, $y\in\mathcal{Y}$ is assigned to a selection likelihood $\pi_t(\mathbf{x})$ during training iteration $t$, which is updated such that:
\begin{equation}
    \pi_{t+1}(\mathbf{x}) = \pi_t(\mathbf{x}) + \lambda\left\| g_{\mathbf{V}_t}\left(\mathbf{f}_{\mathbf{W}_t}(\mathbf{x})\right) - y \right\|_1 ,\,\,\,t=1,\ldots,T
\label{eq:3}
\end{equation}
where $\mathbf{f}_{\mathbf{W}_t}$ and $g_{\mathbf{V}_t}$ are the transformations learned by iteration $t$, $\lambda$ is a constant used to control the update rate of the sample selection likelihood, and $\|\cdot\|_1$ is the $l1$-norm. 
We initialized $\pi_{1}(\mathbf{x})$ to 1 for every sample $\mathbf{x}$. The likelihood is proportional to the probability that a data sample is being selected. The adaptive sample pair formation process is outlined in Algorithm~\ref{Alg:1}. We generate pairs of samples within a speaker $s$ that either belong to the same or different emotions, each of the two cases occurring with probability equal to $0.5$. We select a sample $\mathbf{x}\in \mathcal{X}_{s,c}$ uttered by speaker $s$ with emotion $c$ based on the following probability:
\begin{equation}
P(\mathbf{x})=\frac{\pi_{t}(\mathbf{x})}{\sum_{\mathbf{x_i}\in \mathcal{X}_{s,c}}(\pi_{t}(\mathbf{x_i}))}
\label{eq:4}
\end{equation}
The proposed update in (\ref{eq:3}) renders the pair formation adaptive in two ways: (1) The selection probability of consistently correctly classified samples decreases over time; and (2) The selection probability of misclassified samples increases proportionately to the error of the current iteration.

\begin{algorithm}[t]
\begin{algorithmic}[1]
\STATE Randomly select speaker $s$ and emotion $c_k$
\STATE Generate $a$ from uniform distribution: $a\sim U(0,1)$
\IF {$a>0.5$}
    \STATE Form a pair of the same emotion: Set $c_m = c_k$
    \ELSE
     \STATE Form a pair of different emotion: Randomly select emotion $c_m \neq c_k$
\ENDIF
\STATE Select sample $\mathbf{x_i} \in X_{s,c_k}$ based on (\ref{eq:4})
\STATE Select sample $\mathbf{x_j} \in X_{s,c_m}$ based on (\ref{eq:4})
\STATE Form pair $(\mathbf{x_i},\mathbf{x_j})$
\end{algorithmic}
\caption{Adaptive sample pair formation \label{Alg:1}}\vspace{-5pt}
\end{algorithm}
\setlength{\textfloatsep}{5pt}

\vspace{-8pt}
\subsection{Experimental Design}
\label{sec:ExpDesign}
\vspace{-5pt}
Here, we describe the experimental design that was employed to validate the proposed approach.

\subsubsection{In-domain and out-of-domain learning}
\label{subsubsec:InOutDomain}
\vspace{-10pt}
In order to understand the inherent domain difference between the source and the target, we conduct two experiments without transfer learning. First, we conduct out-of-domain learning by training an emotion classification model on the source data and testing on the target. All samples from the source are used to train the model and all samples from the target are used for testing. Second, we perform in-domain-training, according to which data from the target domain are used for training and testing the models. In-domain-training is evaluated using leave-one-subject-out cross-validation on the target data only. For both in-domain and out-of-domain experiments, we employed a 5-layer FNN with 64, 32, 16, and 16 nodes in the first, second, third, and fourth layers, as well as 3 nodes in the output layer. The rectified linear unit (ReLU) activation function is employed in the first three layers, while the sigmoid activation is used in the last decision making layer to yield a probability outcome for each emotional class. All models are trained to minimize the cross-entropy loss using the Adam optimization~\cite{Kingma2015AdamAM} with a learning rate of 0.0005, 250 epochs.

\subsubsection{Few-shot learning}
\label{subsubsec:FewShot}
\vspace{-5pt}
We examine the effectiveness of the proposed metric learning (i.e., MeL, MeL-S, MeL-ASPF; Sections~\ref{Sec:Siamese}-\ref{Sec:Siamese&supervised&adaptive}) by gradually rendering available an increasing number of labelled samples from the target data in our experiments. We start by randomly selecting 1 labelled sample per speaker and per emotion from the target data and using the remaining samples in the test set. We repeat this process 10 times and compute the average accuracy in order to obtain an unbiased result. Then, we gradually increase the number of samples to $2,\ldots,10$ clips per speaker and per emotion, to explore changes in performance when additional labeled data from the target domain is available.

The proposed MeL, MeL-S, and MeL-S-ASPF approaches employ a SNN, which is first trained on the source data and then fine-tuned on the labelled samples from the target data.

To be consistent with the in-domain and out-of-domain learning (Section~\ref{subsubsec:InOutDomain}), the proposed MeL, MeL-S, and MeL-S-ASPF approaches are implemented with a 5-layer SNN with 64, 32, 16, and 16 nodes and ReLU activation in the hidden layers, as well as a sigmoid activation in the single node of the output layer. Learning is performed using cross-entropy loss and Adam optimization with a learning rate of 0.0005 and 250 training epochs. The trade-off between penalizing dissimilarity between samples belonging to the same class against similarity between samples belonging to different classes, as shown in~(\ref{eq:1}), is set to $\kappa=1$. The supervision layer of the MeL-S model includes 1 hidden layer with 8 units and ReLU activation, and an output layer with 3 nodes and sigmoid activation. The adaptive sample pair formation in MeL-S-ASPF is performed with an update rate $\lambda=0.1$ of the selection likelihood in (\ref{eq:3}), since our goal is to avoid an uninformed abrupt increase of the selection likelihoods in the first few training iterations. The MeL-S-ASPF approach further includes $T=25$ iterations, during which the SNN is trained using 10 epochs. When training the model on the source data, we form pairs of samples without considering speaker identity, which allows us to maximize the generalization of the original model as well as to mitigate the difference between the source datasets. When fine-tuning on the target data, we form pairs of samples within a speaker (i.e., a pair containing samples from the same speaker) with a selection likelihood that follows a uniform distribution with half of the pairs being from the same emotion, while the other half from different emotions. Sample pair formation on the target data using random pairs of speakers was also performed, but provided slightly decreased performance.


\vspace{-8pt}
\subsubsection{Baseline methods}
\label{subsubsec:Baseline}
\vspace{-5pt}
We used two baseline methods to compare the proposed metric learning approaches. The first was implemented through FNN fine-tuning, a commonly used transfer learning technique~\cite{zhang2017speech,feng2020transfer}. Fine-tuning is conducted by pre-training the FNN on the source samples and then refining the learned weights of all layers based on the available labelled target data. We employ 5-layer FNN, similar to Section~\ref{subsubsec:InOutDomain}. The second baseline follows the Adversarial Discriminative Domain Adaptation (ADDA) method~\cite{tzeng2017adversarial}, a recently proposed domain adaptation method, that aims to reduce the distribution difference between the source and target samples through an adversarial loss. We implemented the ADDA in a similar way to previous research on speech emotion recognition~\cite{gideon2019improving} using the time-frequency log-scaled spectrogram patches as features. The 256 dimensional Mel Filter Banks (MFBs) are computed for each sample, with a frame length of 32ms and a 16ms shift using Scipy \cite{virtanen2020scipy}. Each patch is comprised of 30 frames, with an overlap of 15 frames, resulting in a $256\times30$ input to a convolutional neural network (CNN). The CNN uses a kernel size of 15 for the first convolutional layer, and a kernel size of 3 with a dilation rate of 2 for the second convolutional layer. Each convolutional layer has 128 channels, followed by a max pooling layer of pool size 3 and stride size 2. Then a global maximum is computed and resulted in a 128-dimensional feature. Two hidden layers with 128 units were applied to the extracted feature before the output softmax layer. Each layer used the ReLU activation except the output layer, which used sigmoid. Because ADDA does not require target labels during training, we fine-tuned the trained model using target data for a better comparison with other methods. The dropout rate during training was 0.2. The prediction for each audio sample is obtained through a majority voting based on all the patches of the corresponding sample.

\vspace{-5pt}
\section{Results}
\vspace{-5pt}

\begin{figure*}[!htb]
  \centering
  \begin{minipage}{1\linewidth}
  \includegraphics[scale=0.55]{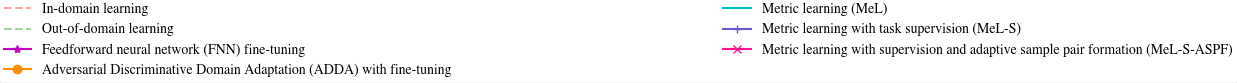}
  \end{minipage}
  \begin{minipage}{0.495\linewidth}
  \centering\includegraphics[trim = 3.4cm 0.7cm 4cm 2.2cm, clip=true, scale=0.273]{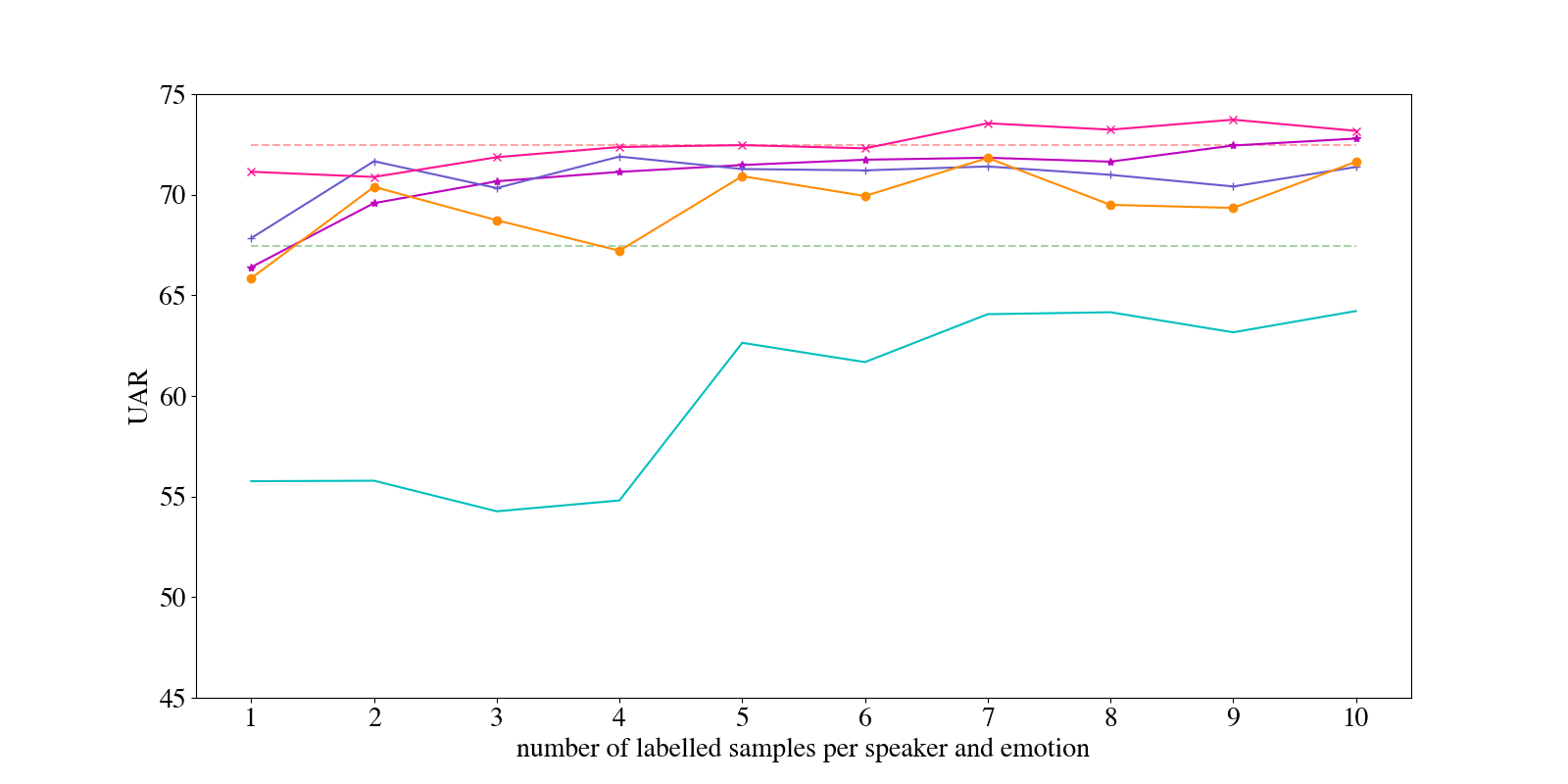}
  \centerline{(a) IEMOCAP-SC}
  \end{minipage}
  \begin{minipage}{0.495\linewidth}
  \centering\includegraphics[trim = 3.4cm 0.7cm 4cm 2.2cm, clip=true, scale=0.273]{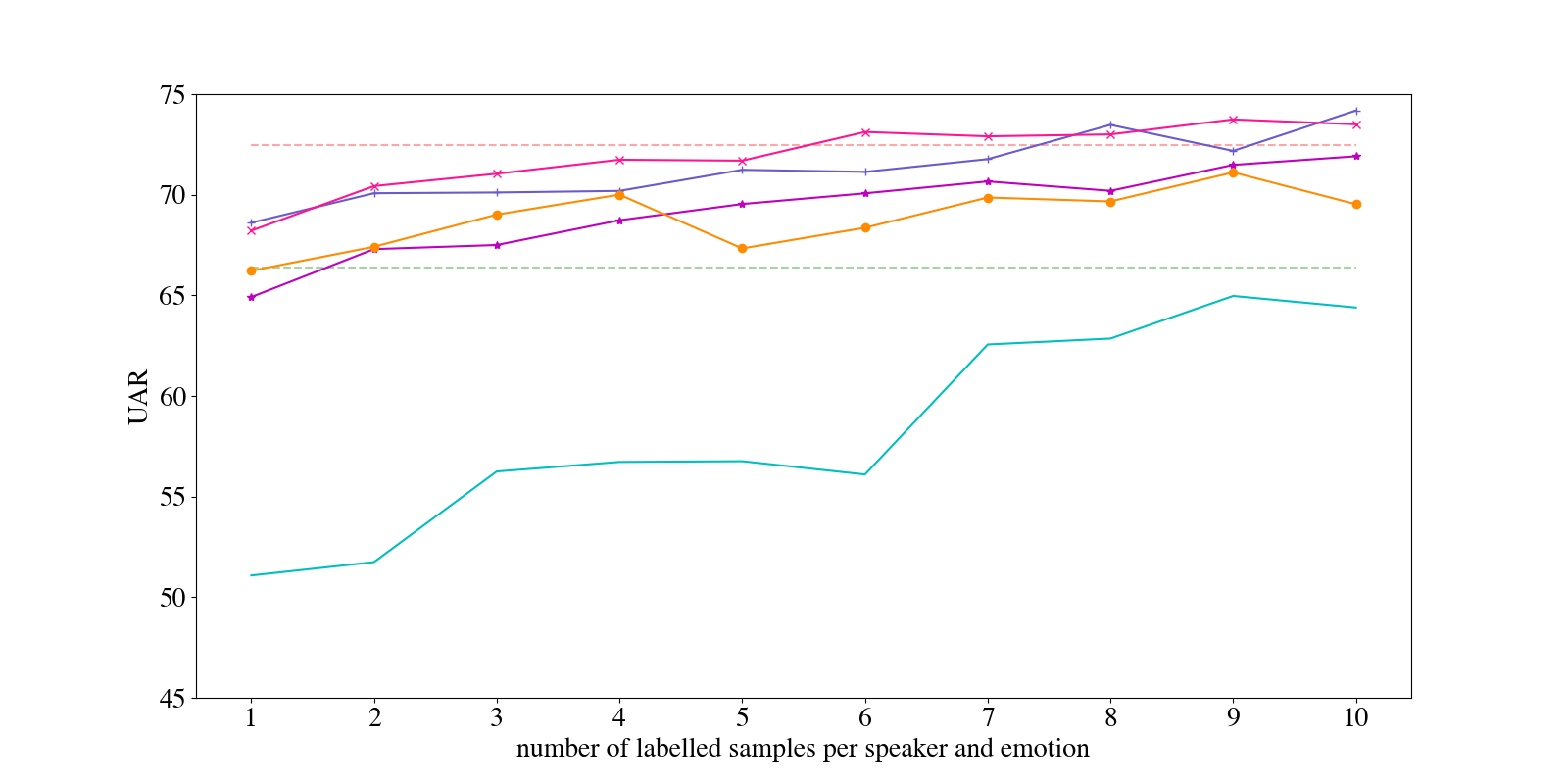}
  \centerline{(b) RAVDESS}
  \end{minipage}
  \begin{minipage}{0.495\linewidth}
  \centering\includegraphics[trim = 3.4cm 0.7cm 4cm 2.2cm, clip=true, scale=0.273]{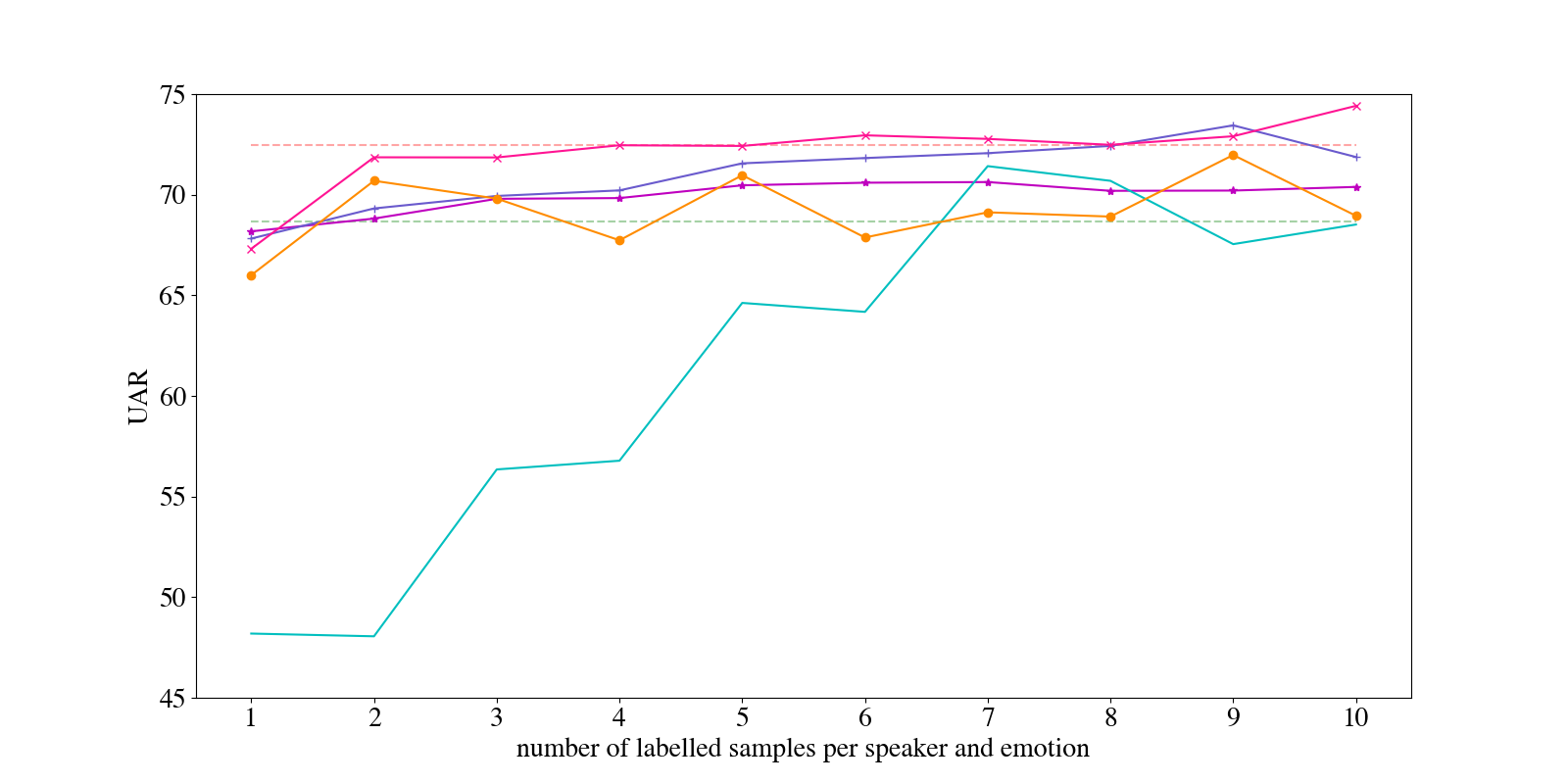}
  \centerline{(c) CREMA-D}\medskip
  \end{minipage}
  \begin{minipage}{0.495\linewidth}
 \centering\includegraphics[trim = 3.4cm 0.7cm 4cm 2.2cm, clip=true, scale=0.273]{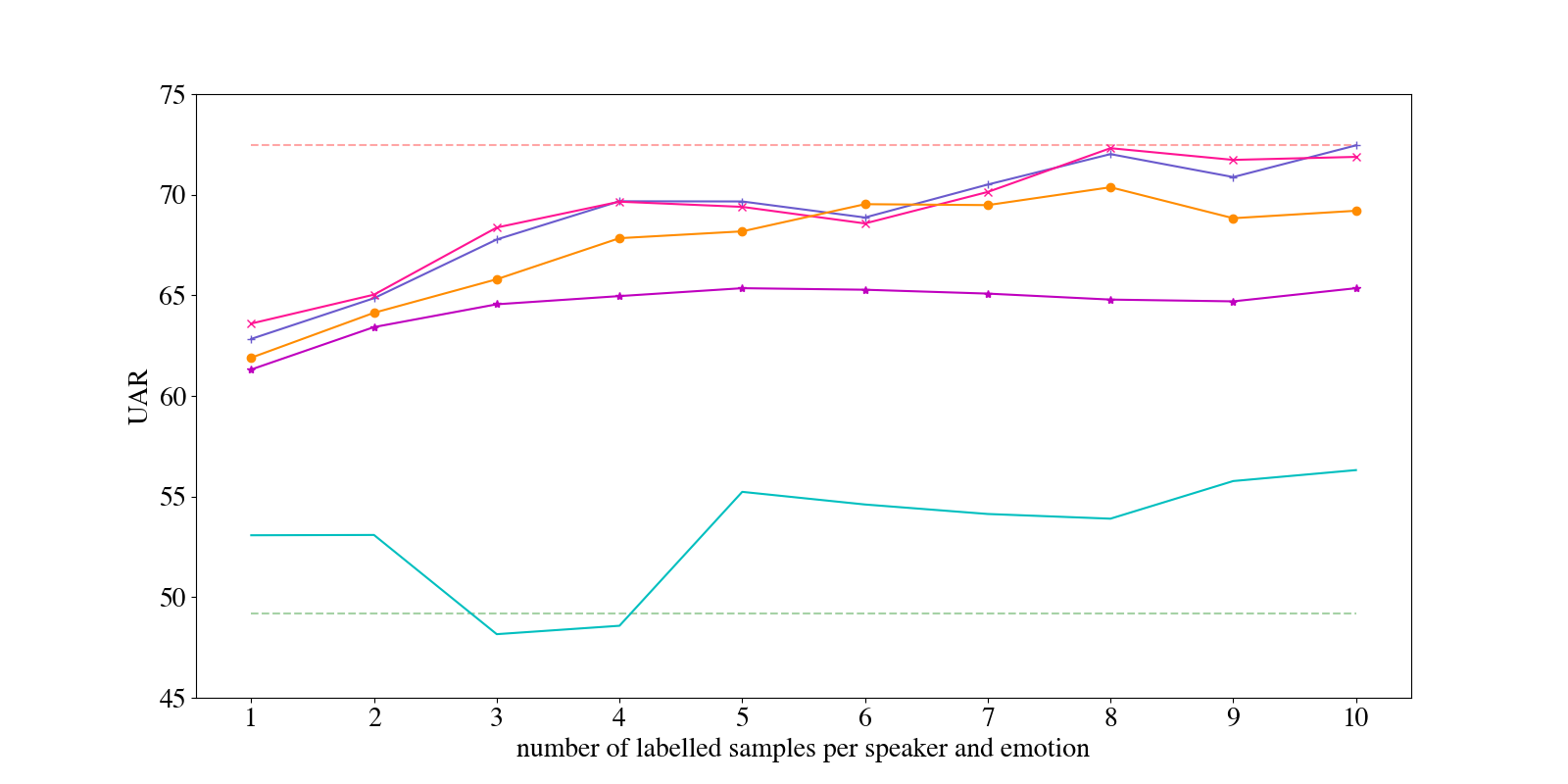}
  \centerline{(d) eNTERFACE’05}\medskip
  \end{minipage}
  \hfill
  \hfill
 \caption{Unweighted average recall (UAR) for in-domain, out-of-domain, and few shot learning methods using various source datasets and an increasing number of labelled samples per speaker and emotion.}
 \vspace{-3pt}
\label{fig:result}
\end{figure*}

All approaches are evaluated on their ability to correctly classify the three considered emotions using the unweighted average recall (UAR), which is computed by taking the average of recall rates for each emotion (Figure~\ref{fig:result}). The chance UAR for the 3-way classification task is 33.3\%. The in-domain training achieves UAR of 72.5\% suggesting moderate to high differentiation between the three emotions when a significant number of target data is available. The out-of-domain performance ranges between 49\% to 68\% demonstrating the potential data mismatch between scripted and spontaneous speech. RAVDESS, CREMA-D, and IEMOCAP-SC provide relatively successful out-of-domain training (i.e., 67.4\%), while this is not the case for eNTERFACE'05, which appears to depict a large domain difference with the target IEMOCAP-SP dataset. This differential result might be attributed to the fact that eNTERFACE'05 was the only dataset in which audio samples were recorded by regular participants rather than actors.

In regards to the proposed metric learning, the MeL approach does not yield satisfactory performance. However, the supervision added as part of the MeL-S and MeL-S-ASPF approaches significantly improves results. A potential reason for this is that MeL-S and MeL-S-ASPF explicitly model the emotional outcome of interest, which is not the case for MeL. The MeL does not outperform the FNN fine-tuning and ADDA potentially for the same reason. It is noteworthy that the MeL-S and MeL-S-ASPF methods depict good performance even when a small amount of labelled target data is available (e.g., 1-2 labelled samples per speaker and emotion for eNTERFACE'05, RAVDESS, and IEMOCAP-SC). Also the MeL-S and MeL-S-ASPF models appear to reach stability in their performance when increasing the number of labelled target data from the target domain. We further observe that the MeL-S-ASPF outperforms the MeL-S across many cases including the IEMOCAP, RAVDESS, and CREMA-D datasets, suggesting that the adaptive sample pair formation benefits performance compared to the random formation of sample pairs. This indicates that this adaptive process can potentially contribute to effectively learning challenging samples that are difficult to estimate when not taking into account the corresponding modeling error. The overall best performance is obtained by the MeL-S-ASPF model when using CREMA-D as the source with 10 samples per speaker and emotion, reaching UAR larger than the in-domain learning (i.e., 74\%).

We further compare the performance of metric learning to the baseline. Both FNN fine-tuning and ADDA appear to be consistently lower compared to the MeL-S and MeL-ASPF. The UAR of FNN fine-tuning, is slowly increasing, as we use more labelled samples from the target data. When we only use one labelled sample per speaker and emotion, negative transfer occurs for the FNN fine-tuning in many cases (e.g., RAVDESS, IEMOCAP-SC), indicating that the corresponding approach may not be promising for transferring knowledge between scripted and spontaneous speech in emotion recognition. The performance increase of ADDA is less consistent compared with FNN fine-tuning, which could possibly be due to unstable training of the adversarial model. However, the best performance of ADDA is better compared to that of the FNN fine-tuning, indicating the potential of this method.



\begin{figure}[!htb]
  \centering
  \begin{minipage}{0.243\textwidth}
  \centering
  \includegraphics[trim = 3.38cm 8.2cm 4cm 8.6cm, clip=true, scale=0.327]{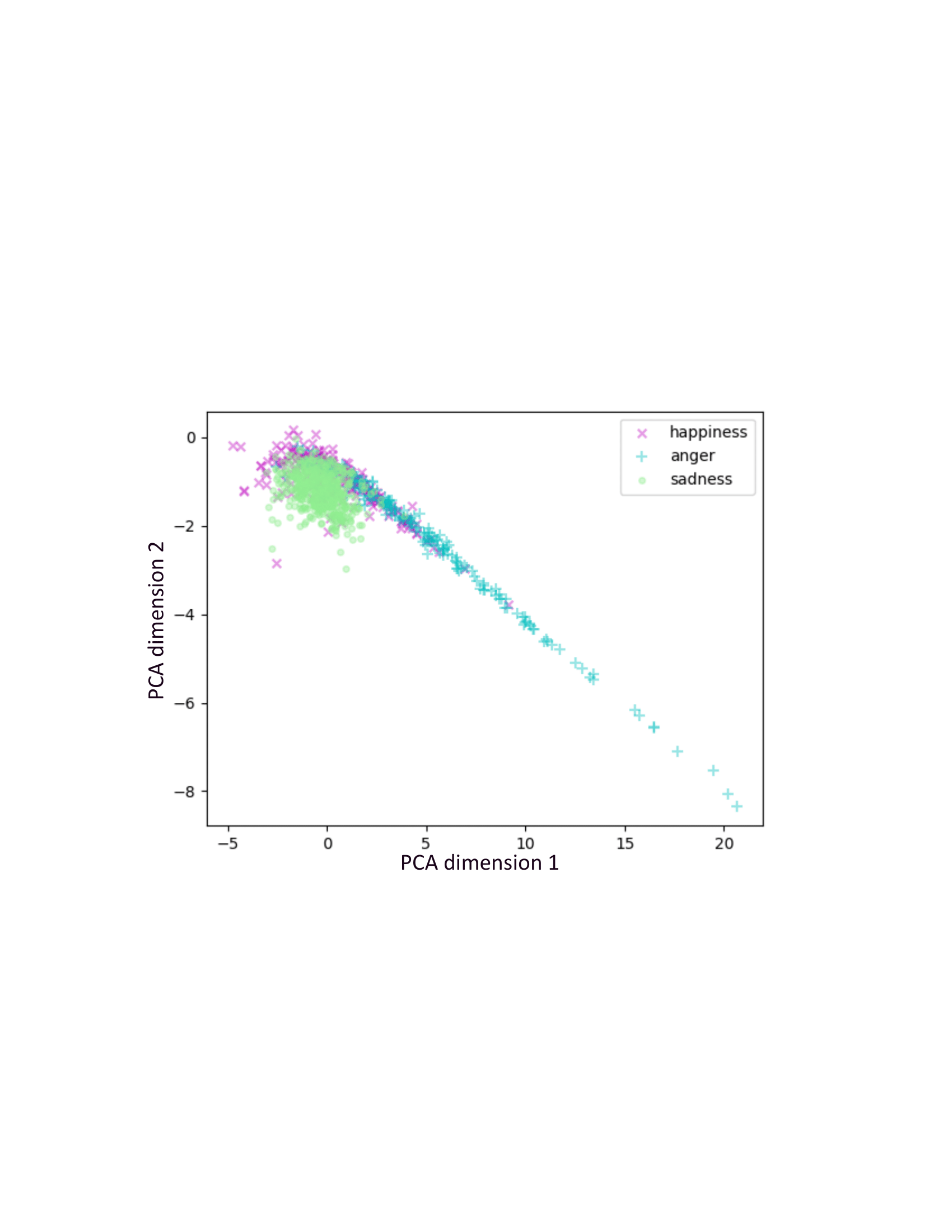}
  \centerline{(a) FNN fine-tuning}
  \end{minipage}
  \begin{minipage}{0.24\textwidth}
  \centering
  \includegraphics[trim = 1.5cm 0cm 1.6cm 0.5cm, clip=true, scale=0.327]{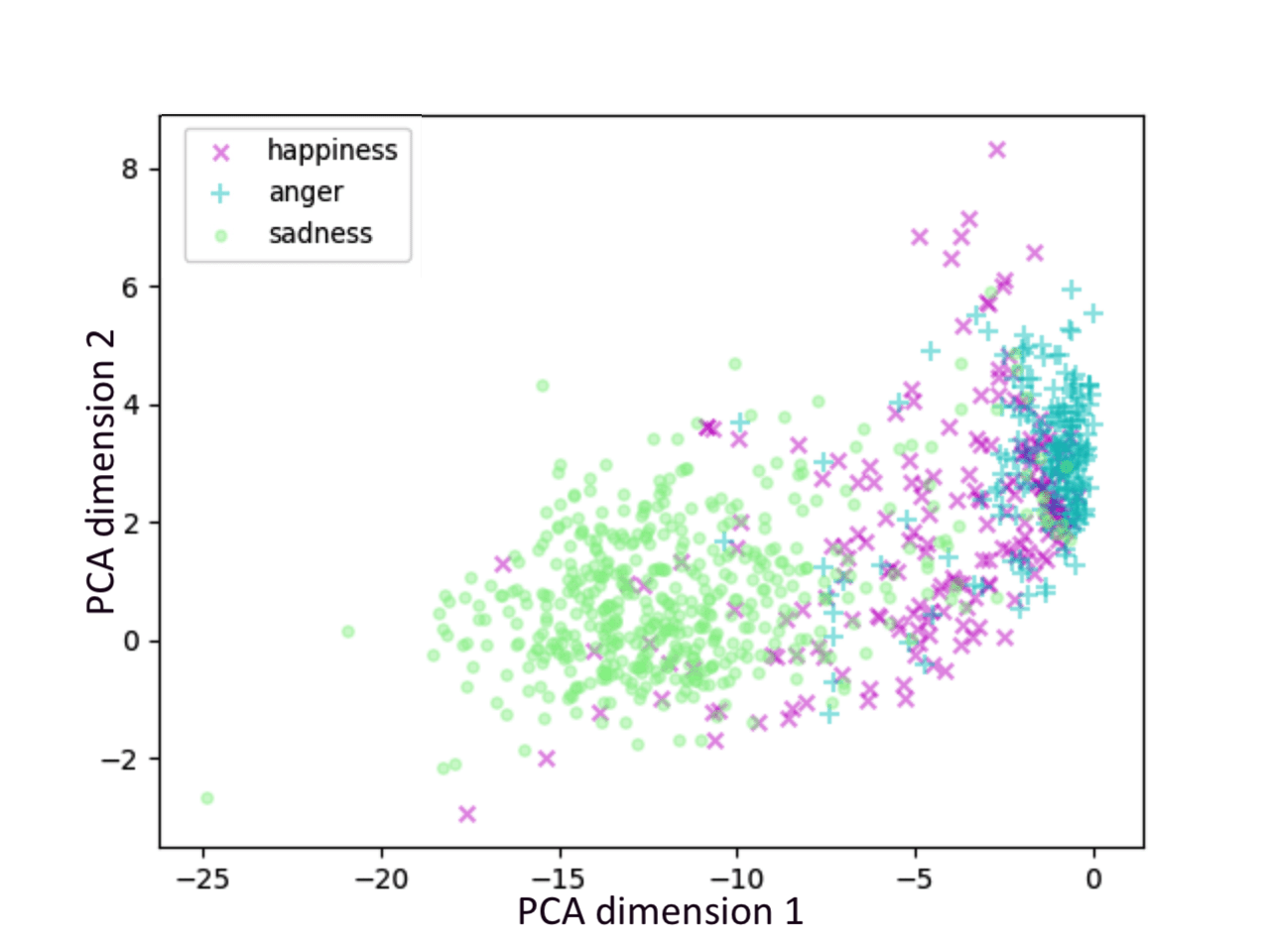}
  \centerline{(b) MeL-S-ASPF}
  \end{minipage}
 \caption{Data distributions learned by the feedforward neural network (FNN) fine-tuning and the metric learning with supervision and adaptive sample pair formation (MeL-S-ASPF) using RAVDESS as the source dataset.}\vspace{-3pt}
\label{fig:pca}
\end{figure}

Next, we visualize the embeddings learned by the proposed metric learning and the baseline non-metric learning approaches. For this, we perform principal component analysis (PCA) to the output of the last hidden layer of each model and provide scatter plots of the first two PCA dimensions corresponding to the largest data variance (Fig.~\ref{fig:pca}). The transformed data samples depict high overlap among the three classes when using the baseline FNN fine-tuning, while the MeL-S-ASPF model results in more distinct distributions. This difference is depicted in our results, since the metric learning approaches with additional supervision outperform the FNN fine-tuning (Fig.~\ref{fig:result}).

\begin{figure*}[t]
  \centering
  \begin{minipage}{0.195\linewidth}
  \includegraphics[trim = 0.6cm 0cm 1cm 1.5cm, clip=true, scale=0.269]{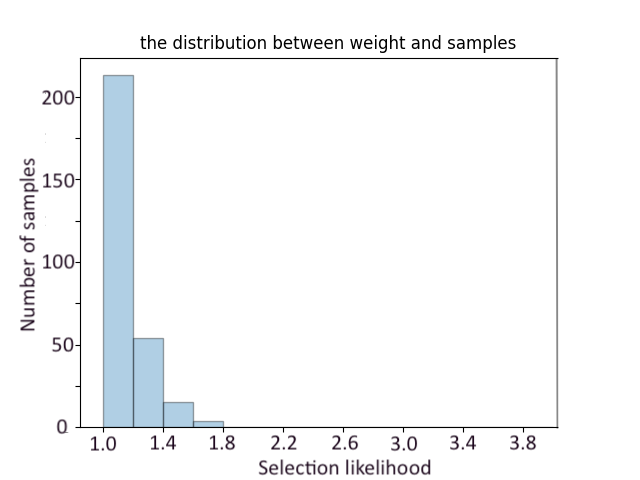}
  \centerline{5th iteration}
  \end{minipage}
  \hfill
  \begin{minipage}{0.195\linewidth}
  \includegraphics[trim = 1.1cm 0cm 1.6cm 1.5cm, clip=true, scale=0.267]{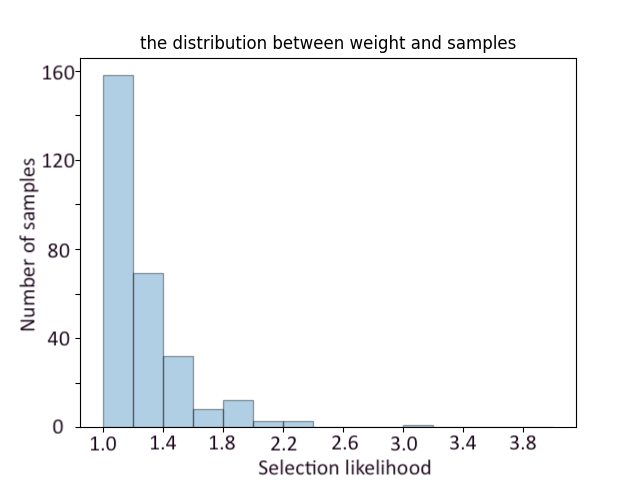}
  \centerline{10th iteration}
  \end{minipage}
  \hfill
  \begin{minipage}{0.195\linewidth}
  \includegraphics[trim = 1.1cm 0cm 1.6cm 1.5cm, clip=true, scale=0.267]{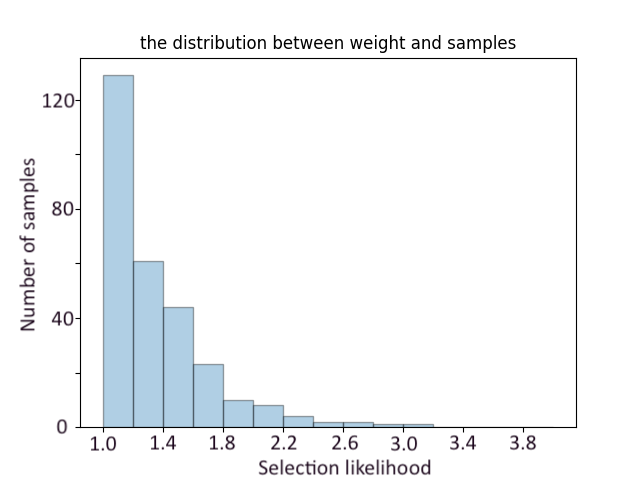}
  \centerline{15th iteration}
  \end{minipage}
  \hfill
  \begin{minipage}{0.195\linewidth}
  \includegraphics[trim = 1.1cm 0cm 1.6cm 1.5cm, clip=true, scale=0.267]{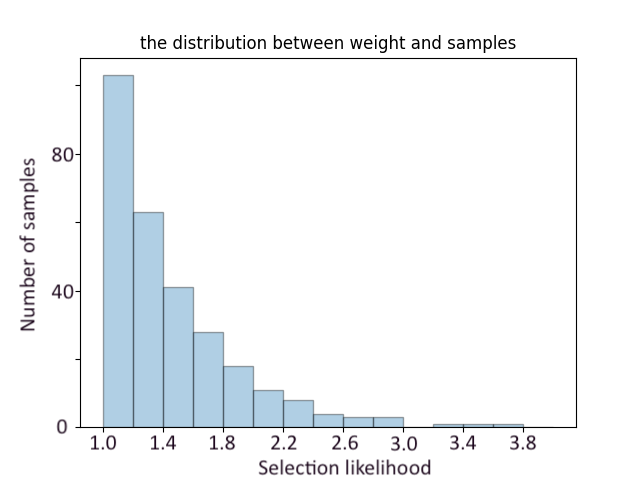}
  \centerline{20th iteration}
  \end{minipage}
  \hfill
  \begin{minipage}{0.195\linewidth}
  \includegraphics[trim = 1.1cm 0cm 1.6cm 1.5cm, clip=true, scale=0.267]{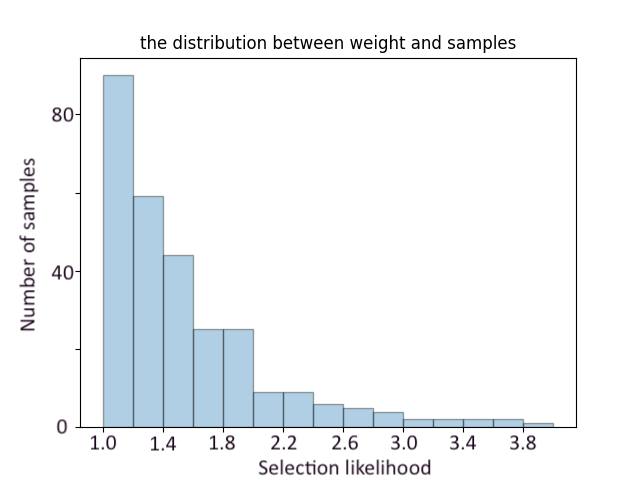}
  \centerline{25th iteration}
  \end{minipage}
  \smallskip
  \vspace{1pt}
\centerline{(a) CREMA-D}
\medskip
  \centering
  \begin{minipage}{0.195\linewidth}
  \includegraphics[trim = 0.6cm 0cm 1.6cm 1.5cm, clip=true, scale=0.269]{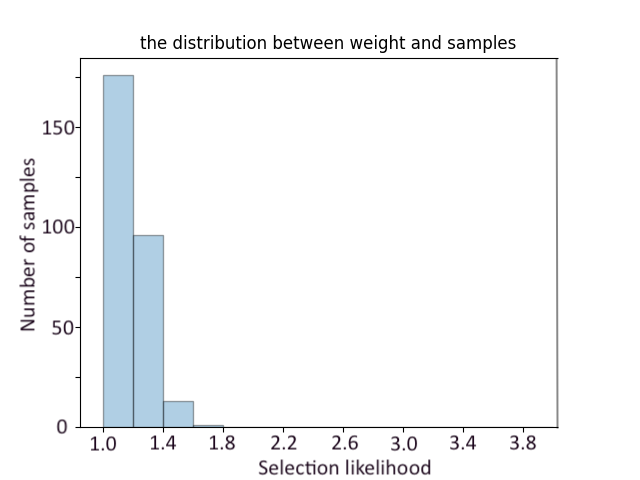}
  \centerline{5th iteration}
  \end{minipage}
  \hfill
  \begin{minipage}{0.195\linewidth}
  \includegraphics[trim = 1.1cm 0cm 1.6cm 1.5cm, clip=true, scale=0.267]{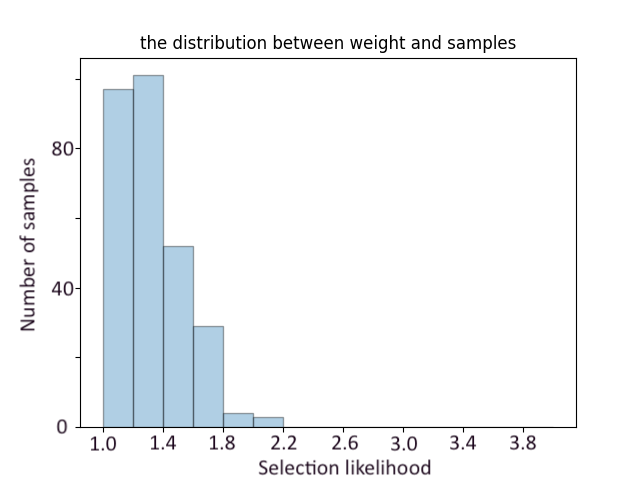}
  \centerline{10th iteration}
  \end{minipage}
  \hfill
  \begin{minipage}{0.195\linewidth}
  \includegraphics[trim = 1.1cm 0cm 1.6cm 1.5cm, clip=true, scale=0.267]{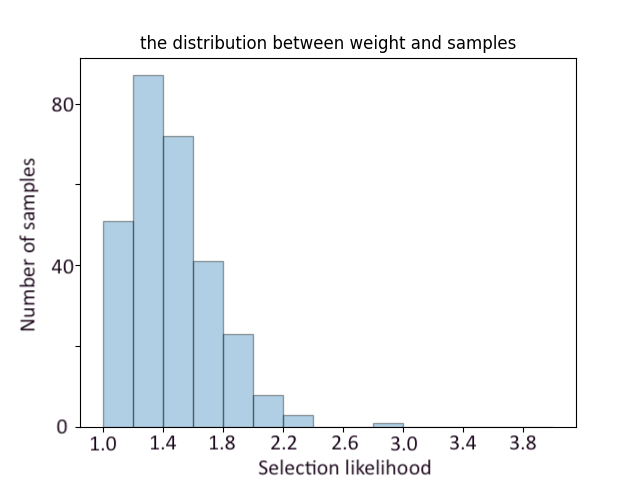}
  \centerline{15th iteration}
  \end{minipage}
  \hfill
  \begin{minipage}{0.195\linewidth}
  \includegraphics[trim = 1.1cm 0cm 1.6cm 1.5cm, clip=true, scale=0.267]{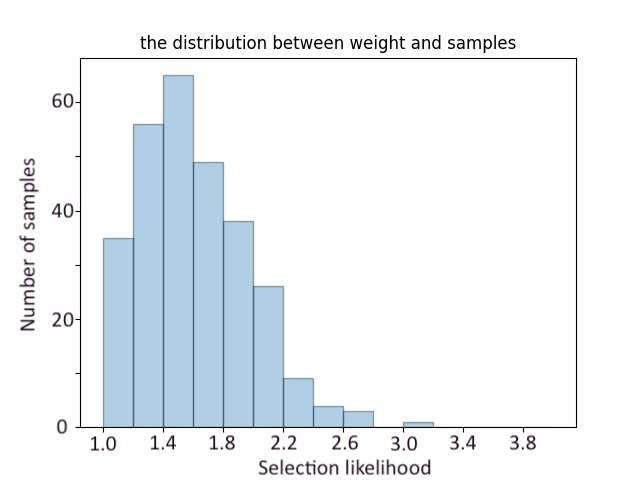}
  \centerline{20th iteration}
  \end{minipage}
  \hfill
  \begin{minipage}{0.195\linewidth}
  \includegraphics[trim = 1.1cm 0cm 1.6cm 1.5cm, clip=true, scale=0.267]{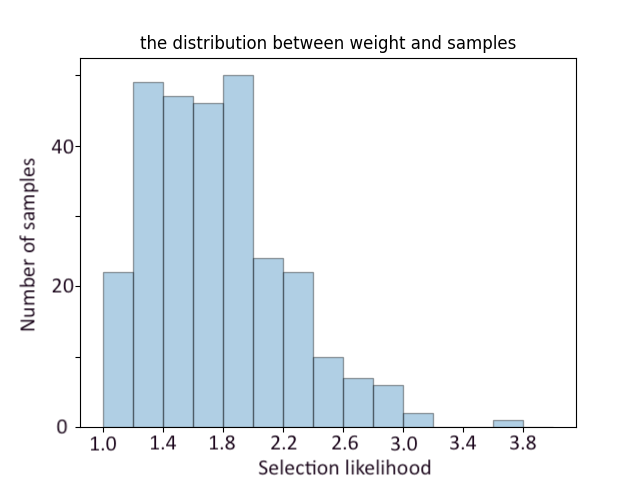}
  \centerline{25th iteration}
  \end{minipage}
  \centerline{(b) RAVDESS}
  \vspace{-10pt}
  \caption{The distribution of sample selection likelihoods $\pi_t(\mathbf{x})$ ($t=5,10,15,20,25$) using the metric learning with supervision and adaptive sample pair formation (MeL-S-ASPF) for CREMA-D and RAVDESS over various training iterations $t$.}\vspace{-5pt}
\label{fig:weight}
\end{figure*}

We investigate how sample selection likelihoods vary throughout the adaptive sample pair formation process by tracking the selection likelihood $\pi_t(\mathbf{x_n})$ of training samples $\mathbf{x_n}$ over different training iterations $t$ ($t=5,10,15,20,25$) (Fig.~\ref{fig:weight}). When using the CREMA-D as the source, the selection likelihood of most samples is concentrated around 1. This suggests that samples from CREMA-D are fairly easily classified based on the MeL-S-ASPF, therefore there is no need to update the selection likelihood for many samples. In contrast, many samples from the RAVDESS dataset are updated with the corresponding selection likelihoods being in higher ranges compared to CREMA-D. This is also depicted in the emotion classification results (Fig.~\ref{fig:result}), where the MeL-S-ASFP presents clear benefits for CREMA-D.


We finally perform error analysis to identify consistently misclassified samples in the target data (independently of the source) and potential sources of error. Common sources of error correspond to noisy audio samples with almost ineligible speech, samples with high discrepancy among annotators, as well as samples for which the linguistic information played an important role in making the final emotion decision. We provide the specific files that correspond to the misclassified samples in Table~\ref{table:Misclassified}.

\begin{table}[t]
\scriptsize
\vspace{-5pt}
\caption{Consistently misclassified samples from target dataset IEMOCAP-SP}
\vspace{-10pt}
\centerline{
\begin{tabular}{lll}
\hline
Ses01M\_impro01\_F001 & Ses02F\_impro03\_F017 & Ses03M\_impro03\_F027\\
Ses05F\_impro03\_F032 & Ses02M\_impro02\_F004 & Ses04F\_impro02\_F009\\
Ses05M\_impro02\_M007 & Ses02M\_impro02\_F004 & \\
\hline
\end{tabular}}
\label{table:Misclassified}
\end{table}

\vspace{5pt}
\section{Discussion \& Conclusions}\vspace{-5pt}
We propose a few-shot learning approach that leverages the abundance of publicly available data from scripted speech in order to detect emotion in spontaneous speech. Our results indicate the feasibility of using scripted speech data to initialize emotion classification models, which can provide useful information for the target data. They further suggest that the proposed MeL-S and MeL-S-ASPF approaches can result in more effective transfer of knowledge compared to conventional distribution learning that models the absolute sample distribution. MeL-S and MeL-S-ASPF yield a relative improvement of 6-23\% compared to out-of-domain learning and 1-7\% compared to the FNN fine-tuning baseline. Indicatively prior work depicts approximate relative improvement of 6-8\%~\cite{zong2016cross,abdelwahab2018domain,gideon2019improving} compared to out-of-domain training for similar emotion classification tasks. For a few cases, the proposed metric learning approach is able to outperform in-domain learning, which suggests that including additional data with the appropriate handling of the potential domain mismatch can benefit learning. Furthermore, this method could potentially be used when detecting different emotions in the source and the target data. Since the SNN learns the relative distance between classes instead of the actual class labels, the learned relations could still provide a meaningful initialization of the weights of the network so that it can be effectively generalized to unseen classes. 

Results from this work can help toward reliable systems of emotional intelligence that can detect emotions from spontaneous speech with a small number of labelled samples from the target domain. This can be particularly useful in real-life applications that rely on tracking well-being from speech, since it demonstrates the ability of speech-based systems to generalize to a new domain using limited supervised experience, potentially mimicking the human ability to recognize emotions from few examples through relative comparison from previous experience. The proposed methods could eventually benefit ambulatory monitoring applications in the clinical domain, where labelled ambulatory data with high temporal granularity are scarce. In this context, audio data collected in the clinic can robustly initialize emotion recognition models that can be generalized in real-life settings for emotion tracking. Since the experience of everyday emotions plays a significant role in psychopathology, this can contribute to effective mental health diagnosis and intervention tracking.

Our study depicts the following limitations. First, the spontaneous speech data from IEMOCAP-SP are collected in laboratory conditions involving high quality microphones and low levels of noise. Despite the phonetic and prosodic differences between scripted and spontaneous speech, the fact that both source and target data have been collected in laboratory conditions has the potential to decrease the domain mismatch between them. As part of our future work, we plan to examine the effectiveness of the proposed approach with various types of real-life datasets, such as the EmotiW~\cite{dhall2019emotiw}. Second, the source and target audio samples are uttered in English, however, emotion is expressed universally, so it would be worthwhile to explore transfer learning and few-shot learning techniques in cross-linguistic scenarios. Of particular interest would be few shot learning between languages of different families, such as between English (i.e., Indo-European language) and Mandarin Chinese (i.e., Sino-Tibetan language family). Datasets that could be used toward this purpose include the Chinese natural emotional audio--visual database (CHEAVD) \cite{li2017cheavd} and the Mandarin Chinese Emotional Speech Dataset - Portrayed (MES-P) \cite{xiao2018mes}. Finally, we studied each of the source datasets in isolation. Inspired by prior findings~\cite{gideon2019improving,li2019exploring}, it would be beneficial to explore the extent to which the combination of multiple source datasets can benefit few-shot learning. It would be also interesting to investigate the interplay between the properties of the various source datasets (e.g., including complementary information) and their effectiveness in few-shot learning.

\section*{Acknowledgment}
Part of this work was funded by the Texas A\&M Data Science Institute (TAMIDS) through the Data Resource Development Program.

\vspace{-10pt}
\bibliographystyle{IEEEtran}
{\linespread{0.93}\selectfont\bibliography{refs}}


\vspace{-30pt}
\begin{IEEEbiography}[{\includegraphics[width=1in,height=1.25in,clip,keepaspectratio]{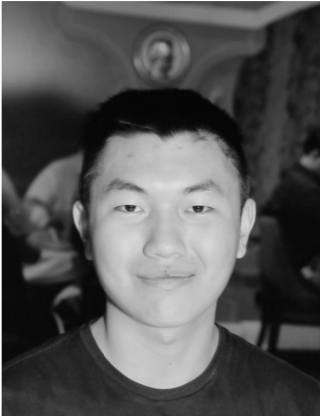}}]
{Kexin Feng} (S'18) received his B.S. degree in Computer Science from Texas A\&M University (2020). He is currently a Ph.D. student in the HUman Bio-Behavioral Signals (HUBBS) Lab at Texas A\&M, and a student member of IEEE and IEEE Signal Processing Society. His research interests lie in machine learning and transfer learning for emotion recognition, as well as its application on social emotion change and individual’s everyday tasks.
\end{IEEEbiography}
\vspace{-30pt}
\begin{IEEEbiography}[{\includegraphics[scale=0.45]{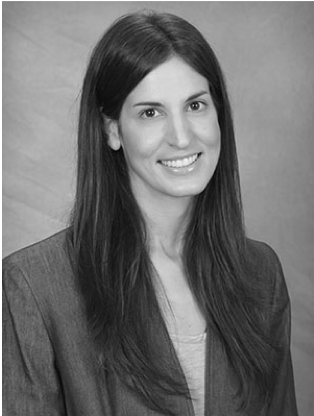}}]{Theodora Chaspari} (S'12, M'17) received her Ph.D (2017) and M.S. (2012) in Electrical Engineering from the University of Southern California, and Diploma in Electrical \& Computer Engineering from the National Technical University of Athens, Greece (2010). She is an Assistant Professor in Computer Science \& Engineering at Texas A\&M and her research interests lie in affective computing and machine learning. She is a recipient of the USC Annenberg Graduate Fellowship 2010, USC Women in Science and Engineering Merit Fellowship 2015, and NSF CAREER Award 2021. Papers co-authored with her students have been nominated and won awards at ACM BuildSys 2019, IEEE ACII 2019, and IEEE BSN 2018.
\end{IEEEbiography}




\end{document}